\documentclass[runningheads]{llncs}
\usepackage[T1]{fontenc}
\usepackage{graphicx}
\usepackage{booktabs}
\usepackage{subcaption}
\usepackage{amsmath, amsfonts}
\usepackage[pagebackref,breaklinks,colorlinks]{hyperref}
\usepackage{tikz}
\usetikzlibrary{positioning, arrows}

\usepackage[textsize=tiny]{todonotes}
\setuptodonotes{inline}

\usepackage[utf8]{inputenc}
\usepackage{xspace}

\usepackage{nicefrac}

\newcommand{\relu}{\text{ReLU}\xspace}
\newcommand{\bn}{\text{BN}\xspace}
\newcommand{\id}{\text{Id}\xspace}
\newcommand{\conv}{\text{Conv}\xspace}
\newcommand{\out}{\text{output}\xspace}

\newcommand{\R}{\ensuremath{\mathbb{R}}}

\renewcommand{\vec}{\textbf}

\mathcode`l="8000
\begingroup
\lccode`~=`l
\lowercase{\endgroup\let~}\ell

\begin{document}
\title{GloNets: Globally Connected Neural Networks}

%
%
\author{Antonio Di Cecco\inst{1}\orcidID{0000-0002-9070-4663} \and
Carlo Metta\inst{2}\orcidID{0000-0002-9325-8232} \and
Marco Fantozzi\inst{3}\orcidID{0000-0002-0708-5495} \and
Francesco Morandin\inst{3}\orcidID{0000-0002-2022-2300} \and
Maurizio Parton\inst{1}\orcidID{0000-0003-4905-3544}}
%
%
\institute{University of Chieti-Pescara, Italy \and
ISTI-CNR, Pisa, Italy \and
University of Parma, Italy}
\maketitle              
\begin{abstract}

Deep learning architectures suffer from depth-related performance degradation, limiting the effective depth of neural networks. Approaches like ResNet are able to mitigate this, but they do not completely eliminate the problem. We introduce Globally Connected Neural Networks (GloNet), a novel architecture overcoming depth-related issues, designed to be superimposed on any model, enhancing its depth without increasing complexity or reducing performance. With GloNet, the network's head uniformly receives information from all parts of the network, regardless of their level of abstraction.
This enables GloNet to self-regulate information flow during training, reducing the influence of less effective deeper layers, and allowing for stable training irrespective of network depth.
This paper details GloNet's design, its theoretical basis, and a comparison with existing similar architectures.
Experiments show GloNet's self-regulation ability and resilience to depth-related learning challenges, like performance degradation. Our findings suggest GloNet as a strong alternative to traditional architectures like ResNets.

\end{abstract}

\section{Introduction}
\label{sec:introduction}

Deep learning's success in AI is largely due to its hierarchical representation of data, with initial layers learning simple features and deeper ones learning more complex, nonlinear transformations of these features \cite{BCVRLR}.
Increasing depth should enhance learning, but sometimes it leads to performance issues \cite{BSFLLT,GlBUDT}. Techniques like normalized initialization \cite{LBOEfB,GlBUDT,SMGESN,HZRDDR} and normalization layers \cite{IoSBNA,BKHLaN} enable up to 30-layer deep networks, but performance degradation persists at greater depths without skip connections.
This issue, detailed in the original ResNet paper \cite{HZRDRL}, stems from the fact that learning identity maps is not easy for a deeply nonlinear layer. ResNet idea is to focus on learning nonlinear ``residual'' information, with a backbone carrying the identity map. This brilliant solution has been key in training extremely deep networks that, when weight-sharing and batch normalization are used, can scale up to thousands of layers.

Deeper neural networks should not experience performance degradation. Theoretically, a deeper network could match the performance of an $n$-layer network by similarly learning features $\mathcal{G}_1, \dots, \mathcal{G}_n$ in its initial layers, then minimizing the impact of additional layers. With this ability to \emph{self-regulate}, such a network could effectively be ``infinitely deep''.
However, even with ResNet architectures, performance degradation persists beyond a certain depth, see Figure~\ref{fig:deep_training_mnist} or~\cite{EbASRN}. This issue can be due to various factors, see for instance~\cite{GlBUDT,SGSTVD,MFPRCUwacv}, and may partly arise from the inability of modern architectures to self-regulate their depth.
Our paper introduces a novel technique to enable self-regulation in neural network architectures, overcoming these depth-related performance challenges.

\textbf{Novel Contributions.}
The main contribution of this paper is introducing and testing GloNet, an explainable-by-design layer that can be superimposed on any neural network architecture, see Section~\ref{sec:model}. GloNet's key feature is its capacity to self-regulate information flow during training. It achieves this by reducing the influence of the deepest layers to a negligible level, thereby making the training more stable, preventing issues like vanishing gradients, and making the network trainable irrespective of its depth, see Section~\ref{sec:experiments}.

This self-regulation capabilities of GloNet lead to several significant benefits:
\begin{enumerate}
    \item \textbf{Faster training:} GloNet trains in half the ResNet time while achieving comparable performance. Beyond the depth threshold where ResNet begins to degrade, GloNet trains in less than half the time and outperforms ResNet.
    \item \textbf{ResNet alternative:} The inability of ResNet-based architectures to self-regulate depth makes GloNet a preferable option, particularly for applications requiring very deep architectures.
    \item \textbf{No NAS needed:} GloNet networks inherently find their effective depth, eliminating the need for computationally expensive Network Architecture Search methods to determine optimal network depth.
    \item \textbf{More controllable efficiency/performance trade-off:} Layers can be selectively discarded to boost efficiency, allowing a controlled trade-off between efficiency and performance, optimizing the network for specific requirements.
\end{enumerate}

\section{Notation and Model Definition}
\label{sec:model}

A feedforward neural network is described iteratively by a sequence of $L$ blocks:
\begin{equation}
    \label{eq:general}
    \vec{x}_{l+1} = \mathcal{G}_l(\vec{x}_l), \quad l=0,\dots,L-1,
\end{equation}
where $\vec{x}_0$ denotes the input vector, and $\vec{x}_{l+1}$ is the output from the $l$-th block. In this context, a ``block'' is a modular network unit, representing a broader concept than a traditional ``layer''. Each block function $\mathcal{G}_l$ typically merges a non-linearity, such as ReLU, with an affine transformation, and may embody more complex structures, like the residual blocks in ResNet.

At the end of the sequence~\eqref{eq:general}, a classification or regression head $\mathcal{H}$ is applied to $\vec{x}_L$. For instance, a convolutional architecture could use a head with average pooling and a fully connected classifier. The fundamental principle in deep learning is that $\mathcal{G}_0,\dots,\mathcal{G}_{l-1}$ hierarchically extract meaningful features from the input $\vec{x}_0$, that can then be leveraged by computing the output of the network:
\[
\out = \mathcal{H}(\vec{x}_L).
\]

When the blocks in~\eqref{eq:general} are simple layers like an affine map followed by a non-linearity (this description comprises, for instance, fully connected and convolutional neural networks), all features extracted at different depths are exposed to the head by a single feature vector $\vec{x}_L$ \emph{that has gone through several non-linearities}. This fact leads to several well-known drawbacks, like vanishing gradients or difficulty in learning when the task requires more direct access to low-level features. When using \relu and shared biases, some low-level information could actually be destroyed. Several excellent solutions have been proposed to these drawbacks, like for instance residual networks~\cite{HZRDRL,HZRIMD}, DenseNets~\cite{HLVDCC}, and preactivated units with non-shared biases~\cite{MFPRCUwacv}.

We propose an alternative solution: a modification to~\eqref{eq:general}, consisting of a simple layer between the feature-extraction sequence and the head, computing the sum of every feature vector.
The architecture is designed to receive information uniformly
from all parts of the network, regardless of their level of abstraction:
\begin{equation}
\label{eq:glonet}
\begin{cases}
\vec{x}_{l+1} = \mathcal{G}_l(\vec{x}_l), \quad l=0,\dots,L-1\\
\vec{x}_{L+1} = \sum_{l=1}^L \vec{x}_l = \sum_{l=0}^{L-1}\mathcal{G}_l(\vec{x}_l)\\
\out = \mathcal{H}(\vec{x}_{L+1})
\end{cases}
\end{equation}

When feature vectors have different dimensions, adaptation to a common dimension is required before the sum, as happens in ResNet. If $\vec{x}_l\in\R^{n_l}$, one can use embeddings in $\R^{\max\{n_l\}}$ to maximally preserve information, and embeddings or projections to $\R^{n_L}$ to maintain the same parameters for the head.
We refer to the additional layer in~\eqref{eq:glonet} as a \emph{GloNet layer}, because all the features $\vec{x}_l$ that without GloNet would be preserved only ``locally'' up to the next $\vec{x}_{l+1} = \mathcal{G}_l(\vec{x}_l)$, appear now in the ``global'' feature vector $\vec{x}_{L+1} = \sum_{l=1}^L \vec{x}_l$ as summands.

\begin{figure}[t]
    \centering
    \includegraphics[scale=0.38]{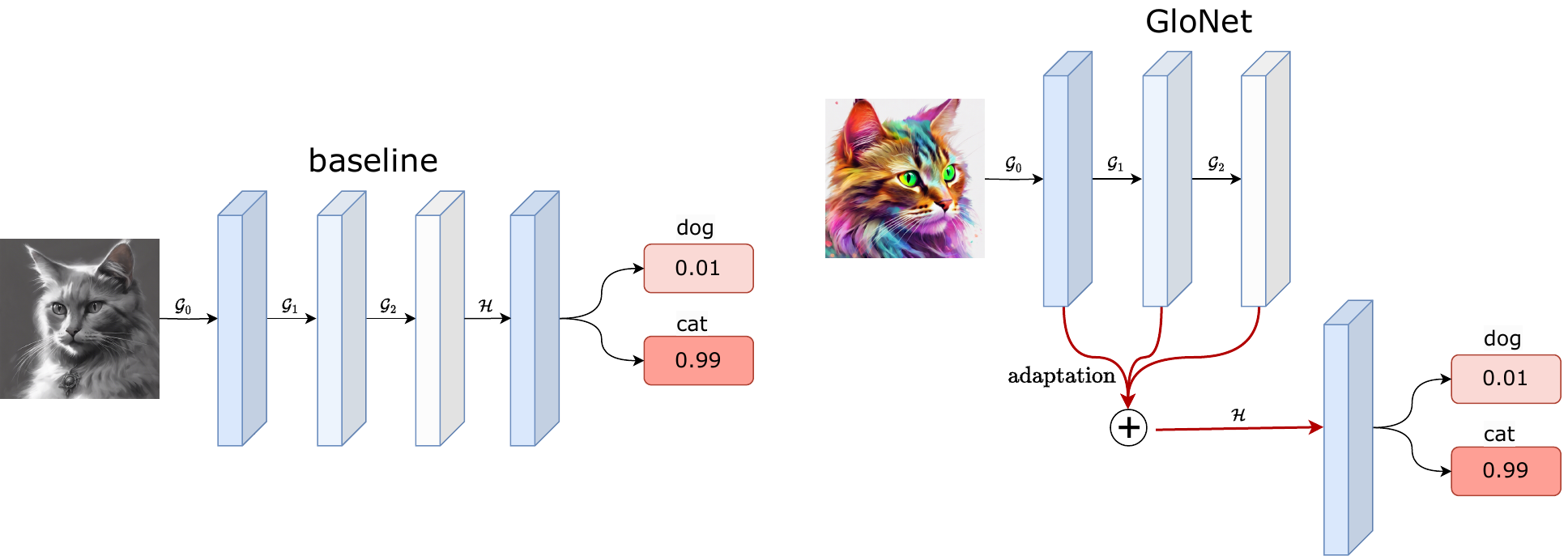}
    \caption{On the left, a 3-blocks neural network followed by a classification head. On the right, the same architecture with GloNet modification in red.
    }
    \label{fig:glonet}
\end{figure}

\begin{remark}[Theoretical support for GloNet self-regulation]
\label{rem:self_regulation}
GloNet provides skip connections solely to the head, and intermediate blocks are not required to learn a residual map, see~\eqref{eq:glonet} and Figure~\ref{fig:glonet}. This ensures direct \emph{and simultaneous} backpropagation pathways from each block, enabling uniform information distribution across the network to the head.
Due to SGD-like training's preference for shorter paths~\cite{ZBHUDL,ZBHUDL2,MaMISR}, \emph{GloNet is expected to accumulate information mostly in the initial blocks rather than the latter ones}, by reducing the influence of the deepest layers to a negligible level. Consequently, GloNet self-regulates its depth during training, rendering it akin to an ``infinitely deep'' architecture. Empirical evidence supporting this claim is presented in Section~\ref{sec:experiments}.
\end{remark}

\begin{remark}[Explainability-by-design]
\label{rem:explainability}
Given the linearity of the GloNet layer, the global feature vector 
$\vec{x}_{L+1}$ provides the contribution that each layer makes to the neural network's prediction. In a feedforward neural network, GloNet enables the analysis of an ensemble of networks represented by the outputs of each network block. This ensemble, comprising blocks $(\mathcal{G}_l)_{l=0,\dots,k}$, functions as individual neural networks, with GloNet integrating their outputs through a linear (currently unweighted) combination.
This architecture allows each sub-network to specialize in learning features at varying levels of granularity, from low-level in early blocks to more complex, large-scale features in later blocks. The linear nature of the GloNet layer facilitates the attribution of importance scores to these features, effectively creating an 'explainable-by-design' tool~\cite{gianfagna2021explainable}.
\end{remark}

\section{Related Work}
\label{sec:related}

In a ResNet with an activation-free backbone, also known as ResNetv2~\cite{HZRIMD}, blocks $\mathcal{G}_l$ are defined as $\id + \mathcal{F}_l$, where $\id$ is the skip connection and $\mathcal{F}_l$ is the ``residual block'', computing two times $\conv\circ\relu\circ\bn$, where \bn is batch normalization. Unrolling the ResNetv2 equation $\vec{x}_{l+1} = \vec{x}_l + \mathcal{F}_l(\vec{x}_l)$ from any block output $\vec{x}_l$ (see~\cite[Equation~4]{HZRIMD}) gives:
\begin{equation}
\label{eq:resnetv2}
\vec{x}_{L} = \vec{x}_l + \sum_{i=l}^{L-1} \mathcal{F}_i(\vec{x}_i)
\end{equation}
This equation shows that ResNet, much like GloNet, passes the output $\vec{x}_l$ of each block directly to the head. However, a key distinction lies in how the head accesses these outputs: ResNet requires distinct pathways for simultaneous access to different outputs, whereas GloNet's head achieves simultaneous access to each output via the GloNet layer. This unique capability of GloNet may contribute to its additional features compared to ResNetv2, as explored in Section~\ref{sec:experiments}.

Moreover, GloNet is faster than ResNet (not requiring batch normalization), and can be seen as an ensemble computing the sum of models of increasing complexity, giving an explainable-by-design model (differently from ResNet).

Unlike DenseNet~\cite{HLVDCC}, aggregating each block with all subsequent blocks through concatenation, GloNet connects only to the last block, with summation for aggregation. This approach avoids the parameter explosion given by concatenation in DenseNet, and maintains the original complexity of the model.

Finally, GloNet can be viewed as a network with early exits at every block, adapted and aggregated before the head. See~\cite{TMKBFI,PSRCDL,GorEar} for the early exit idea.



\section{Implementing GloNet}
\label{sec:implementation}

Implementing GloNet within a certain architecture may not always be as straightforward as described in Section \ref{sec:model}. 
In this Section we describe how GloNet can be implemented in common scenarios.

\textbf{GloNet and Skip Connections.}
If the original architecture includes skip connections (such as ResNet or DenseNet), these should be removed and replaced with the GloNet connection. Otherwise, putting GloNet on top of skip connections, training would not converge as we would be adding the identity to the output multiple times. Note that GloNet provides only skip connections to the GloNet layer, and does not ask the blocks to learn a residual map.

\textbf{GloNet and Batch Normalization.}
Batch normalization plays a major role in enhancing and stabilizing neural network training by normalizing the inputs of each block. Its positive impact is widely acknowledged, though the specific mechanisms of its benefits are still debated \cite{IoSBNA,STIHDB}. Despite these advantages, batch normalization poses challenges, particularly in its interaction with GloNet.
GloNet is designed to dynamically regulate the outputs of different blocks, based on their contribution to the task. It makes negligible the outputs of deeper blocks, a strategy that conflicts with the objectives of batch normalization, which strives to maintain a consistent mean and variance for block inputs.
Consequently, batch normalization, and similarly layer normalization, should be removed prior to the GloNet layer's aggregation. GloNet introduces an alternative form of regularization, which, as we demonstrate in Section~\ref{sec:experiments}, is capable of achieving comparable performances without the need for batch normalization.

\textbf{GloNet into Residual Networks.}
Architectures using residual blocks feature both skip connections and normalization. Once skip connections and normalizations are removed from a ResNetv2 block computing $\id + \text{affine map}\circ\relu\circ\bn\circ\text{affine map}\circ\relu\circ\bn$, one is left with two simpler blocks $\text{affine map}\circ\relu$, and each of those blocks can potentially be aggregated into the GloNet layer. In this case, one ResNetv2 block corresponds to two simpler blocks. This is what we do in this paper, and for this reason when GloNet has $n$ blocks, its equivalent ResNetv2 architecture has $\nicefrac{n}{2}$ blocks.

\textbf{GloNet into Vision Transformers.}
When using more complex architectures like transformers, several different choices can be made, each one potentially affecting the final performance of the GloNet-enhanced model. In this initial exploration of GloNet, we propose a straightforward integration with a Vision Transformer (ViT) \cite{DBKIWW} adapted to CIFAR-10. The image is segmented into 4x4 patches, its class encoded, and then concatenated with the patch encoding and a positional embedding. This series is then fed into a cascade of $n$ encoders with 4 attention heads each, which are accumulated into a GloNet layer, and passed to a classification head. In our experiment, we compared $n=4, 5$ and $6$.





\section{Experiments}
\label{sec:experiments}

In this section we provide experiments supporting the core claims of our paper, as stated in the Introduction. In particular, we focus on showing that GloNet trains much faster than ResNet, that GloNet performances are on par with ResNet's ones, that GloNet can self-regulate its depth, that GloNet does not need batch normalization, and that GloNet is virtually immune to depth-related problems.
All the experiments can be reproduced using the source code provided at~\cite{DiCGlo}.

\textbf{SGEMM Fully Connected Regression.}
We experimented with a regression task from the UCI repository~\cite{sgemm2018,UCIKER}, focused on predicting the execution time of matrix multiplication on an SGEMM GPU kernel. See~\cite{sgemm2018,UCIKER} for details on this task and SGEMM dataset.

Since GloNet has skip connections, to obtain a fair comparison we used a ResNetv2-like baseline. For comparison, we used also a vanilla baseline, identical to the ResNetv2 baseline but without skip connections. Moreover, since GloNet does not use batch normalization (in fact, GloNet self-regulation capabilities can be tampered by normalization, see Section~\ref{sec:implementation}), we also experimented with a vanilla and a ResNetv2 baseline with the \bn layer removed. GloNet and the corresponding baselines (denoted by vanilla, ResNetv2, vanilla-no-\bn, and ResNetv2-no-\bn in figures) have a similar amount of parameters, the only difference being given by the trainable \bn parameters. All blocks have 16 units. All models starts with a linear layer mapping the 14-dimensional input to $\R^{16}$, and ends with the head, a linear layer with 1 unit. GloNet models have an additional GloNet layer before the head \emph{with no additional parameters}.  The number of blocks ranges in [10, 24, 50, 100, 200], respectively (halved for the ResNets because every block is twice the layers of the corresponding non-ResNet model).

For training, we used MSE loss, L$^2$-regularization with a coefficient of $10^{-5}$ (we also tried $10^{-4}$ without improvements), Adam optimizer with a batch size of 1024, learning rate set to $0.01$, He normal initializer for weights, and zero initializer for biases.  We trained all models for 200 epochs, the point at which baselines plateaued, potentially favoring them over GloNet. The first thing to notice is that with GloNet \emph{training takes almost half or less than half the time of ResNet}, see Table~\ref{tab:training_times}. This is because GloNet, differently from ResNet, does not need batch normalization.

\begin{table}[h]
\caption{Average epoch's training time in seconds at different depths, for GloNet and its equivalent ResNetv2 baseline, on SGEMM regression task.}
\label{tab:training_times}
\centering
\begin{tabular}{cccccccc}
\toprule
architecture & \multicolumn{7}{c}{depth} \\
\hline
 & 10    & 24    & 50    & 100   & 200   & 600   & 1000  \\
\hline
ResNetv2       & 42 & 65 & 96 & 147 & 231 & 410 & 675 \\
Glonet          & 27 & 39 & 55 & 86 & 121 & 175 & 289 \\
\bottomrule
\end{tabular}
\end{table}

\begin{figure}[tb]
    \centering
    \includegraphics[width=\textwidth]{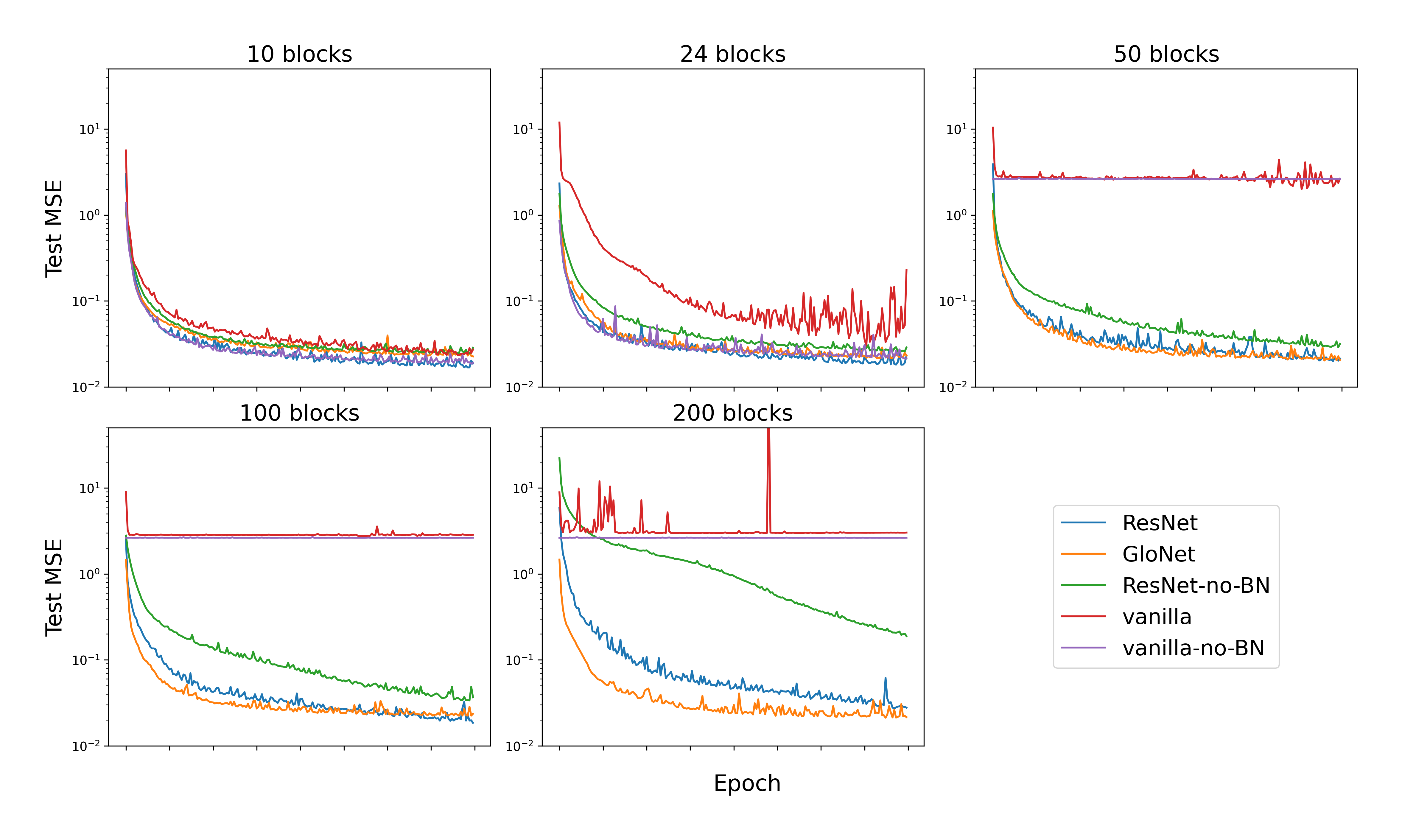}
    \caption{Test errors while training for 200 epochs GloNet and four corresponding baselines at different depths on SGEMM. Starting at 24 blocks, GloNet and ResNetv2 outperform vanilla, as expected due to skip connections. At 50 and 100 blocks, both ResNetv2 and GloNet show depth resilience, with ResNetv2 slightly leading in best test error. ResNetv2-no-BN exhibits significantly higher error, highlighting ResNetv2's reliance on batch normalization. At 100 blocks GloNet curve shows a markedly higher curvature than ResNetv2. At 200 blocks, GloNet outperforms ResNetv2 in best test error and curve shape.}
    \label{fig:sgemm_mse}
\end{figure}

After 200 training epochs, we compared the best test errors and learning curves across different block configurations, see Figure~\ref{fig:sgemm_mse} for learning curves and Table~\ref{tab:best_test_errors} for best test errors. At 200 blocks, GloNet surpassed ResNet in both best test error and learning curve shape. See the caption of Figure~\ref{fig:sgemm_mse} for details on the results of this experiment.

The training shapes in Figure~\ref{fig:sgemm_mse} suggested a unique aspect of GloNet not present in the ResNetv2 baseline. GloNet's training was unaffected by the increasing depth, as shown by the shape of the learning curve that remained consistent whether the network had 10, 24, 50, 100, or 200 blocks. On the contrary, ResNet learning curve became flatter when depth is increasing.

\begin{figure}[ht!]
  \centering
  \includegraphics[width=0.9\columnwidth]{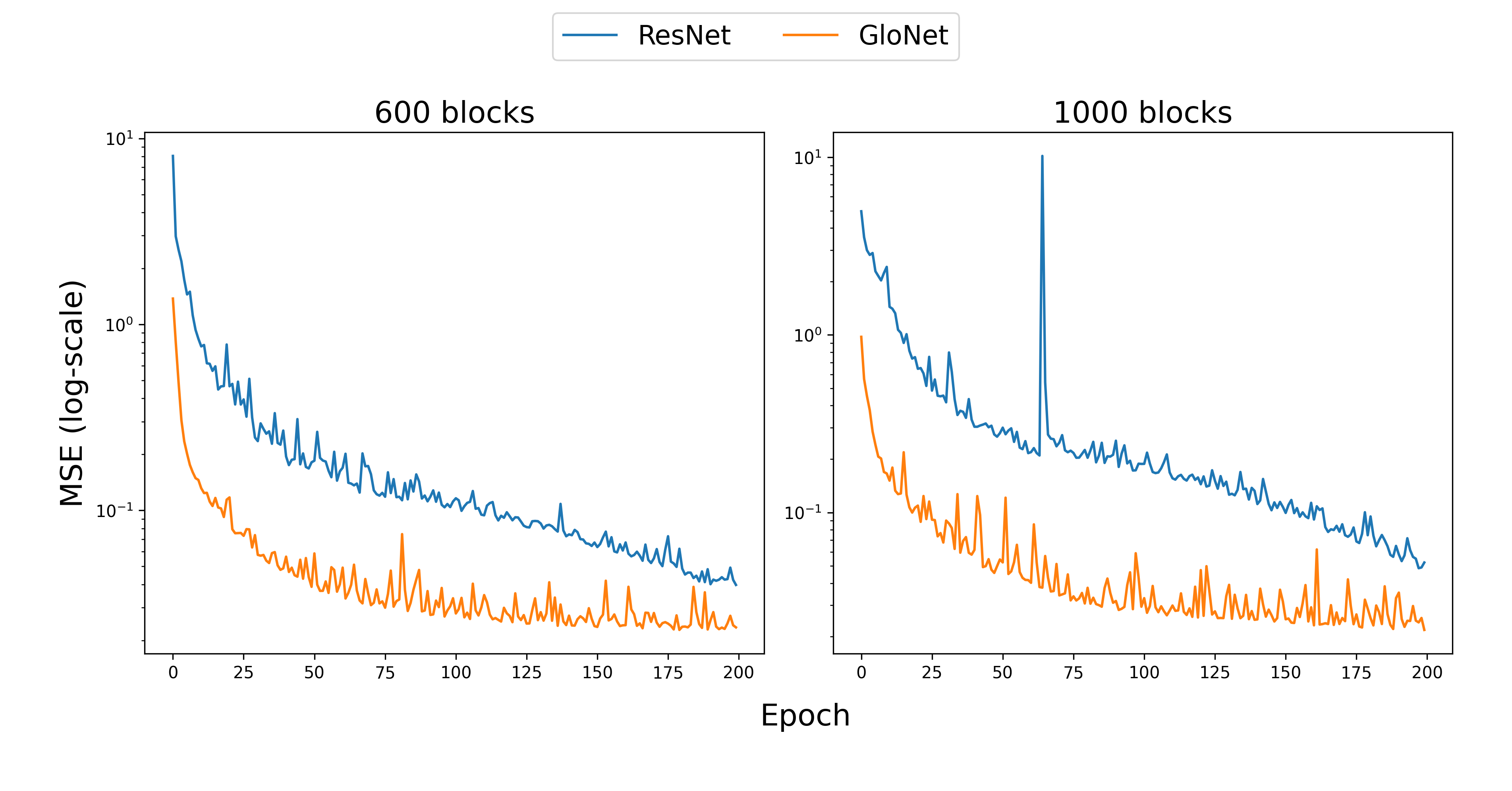}
  \caption{Test MSE learning curves of GloNet (orange) with 600 blocks (left) and 1000 blocks (right), against the equivalent ResNetv2 baseline (blue). Comparing with Figure~\ref{fig:sgemm_mse} shows these substantial depths severely degrade ResNet performance, but do not affect at all GloNet performance.}
  \label{fig:deep_curves}
\end{figure}

To further explore this feature, GloNet was tested with even deeper models (600 and 1000 blocks), and compared against the corresponding ResNetv2 baseline. Even at these substantial depths, GloNet's learning curve maintained its shape, as shown in Figure~\ref{fig:deep_curves}. Moreover, GloNet's performance remained stable across these varying depths, maintaining a best test error of around $0.02$ regardless of the number of blocks ($10, 24, 50, 100, 200, 600$, or 1000), see Table~\ref{tab:best_test_errors}.

\begin{table}[b]
\caption{Best test errors across different network depths.}
\label{tab:best_test_errors}
\centering
\begin{tabular}{lccccccc}
\toprule
\multicolumn{1}{c}{architecture} & \multicolumn{7}{c}{depth} \\
\hline
& 10    & 24    & 50    & 100   & 200   & 600   & 1000  \\
\hline
vanilla         & 0.023 & 0.030 & 1.997 & 2.750 & 2.985 & -     & -     \\
vanilla-no-BN   & 0.018 & 0.021 & 2.624 & 2.624 & 2.624 & -     & -     \\
ResNetv2       & 0.018 & 0.019 & 0.020 & 0.019 & 0.027 & 0.040 & 0.048 \\
ResNetv2-no-BN & 0.024 & 0.026 & 0.029 & 0.033 & 0.189 & -     & -     \\
Glonet          & 0.021 & 0.021 & 0.020 & 0.022 & 0.021 & 0.022 & 0.022 \\
\bottomrule
\end{tabular}
\end{table}

In contrast, ResNet's showed a clear decline as the network depth increased. While its best test error remained around $0.02$ up to 200 blocks, this error increased to $0.04$ and almost $0.05$ at 600 and 1000 blocks, respectively, see again Table~\ref{tab:best_test_errors}. As happened with 100 and 200 blocks, the ResNet learning curve was flatter, diverging from GloNet's more consistent curve shape as depth increased.

\begin{figure}[t!]
  \centering  
  \begin{subfigure}[b]{0.45\textwidth}
    \includegraphics[width=\textwidth]{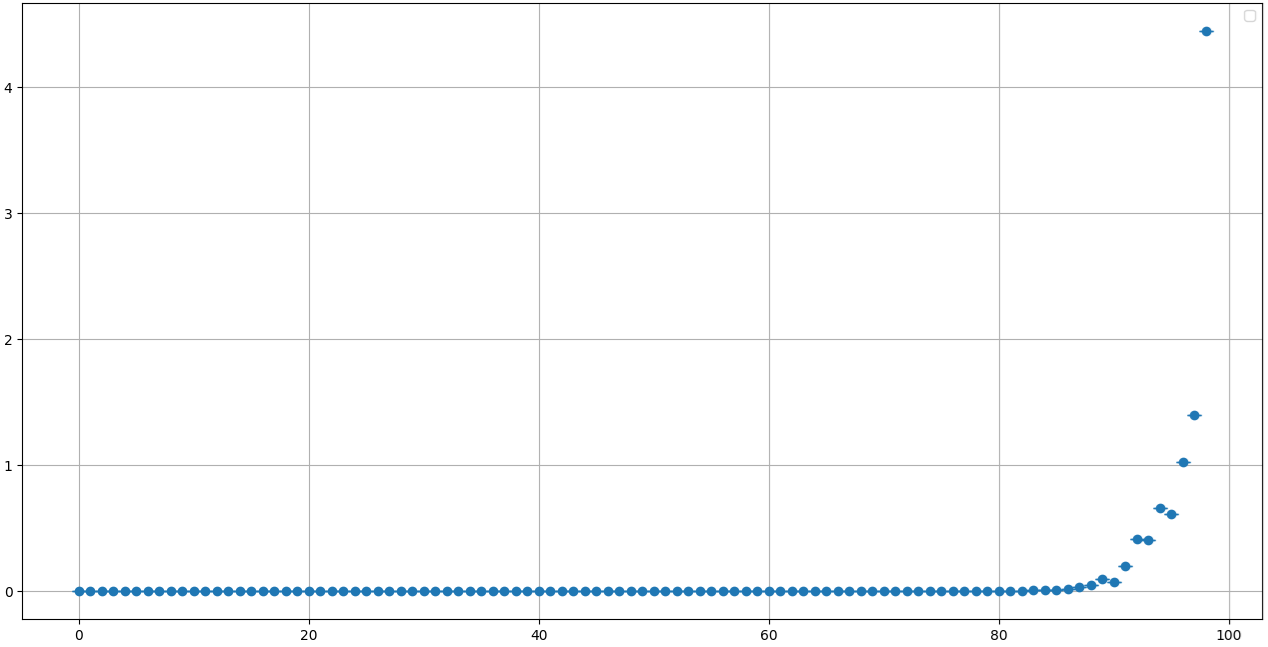}
    \caption{vanilla-no-BN}
    \label{fig:sub1}
  \end{subfigure}
    \hfill 
  \begin{subfigure}[b]{0.45\textwidth}
    \includegraphics[width=\textwidth]{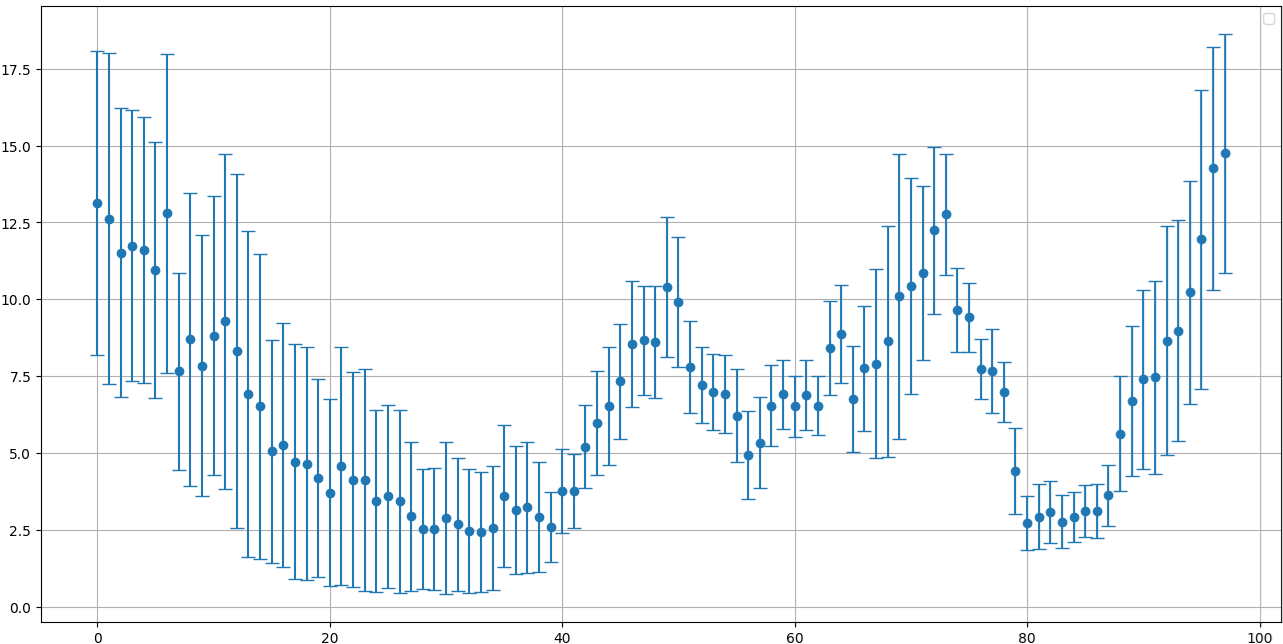}
    \caption{ResNetv2-no-BN}
    \label{fig:sub2}
  \end{subfigure}
  \begin{subfigure}[b]{0.45\textwidth}
    \includegraphics[width=\textwidth]{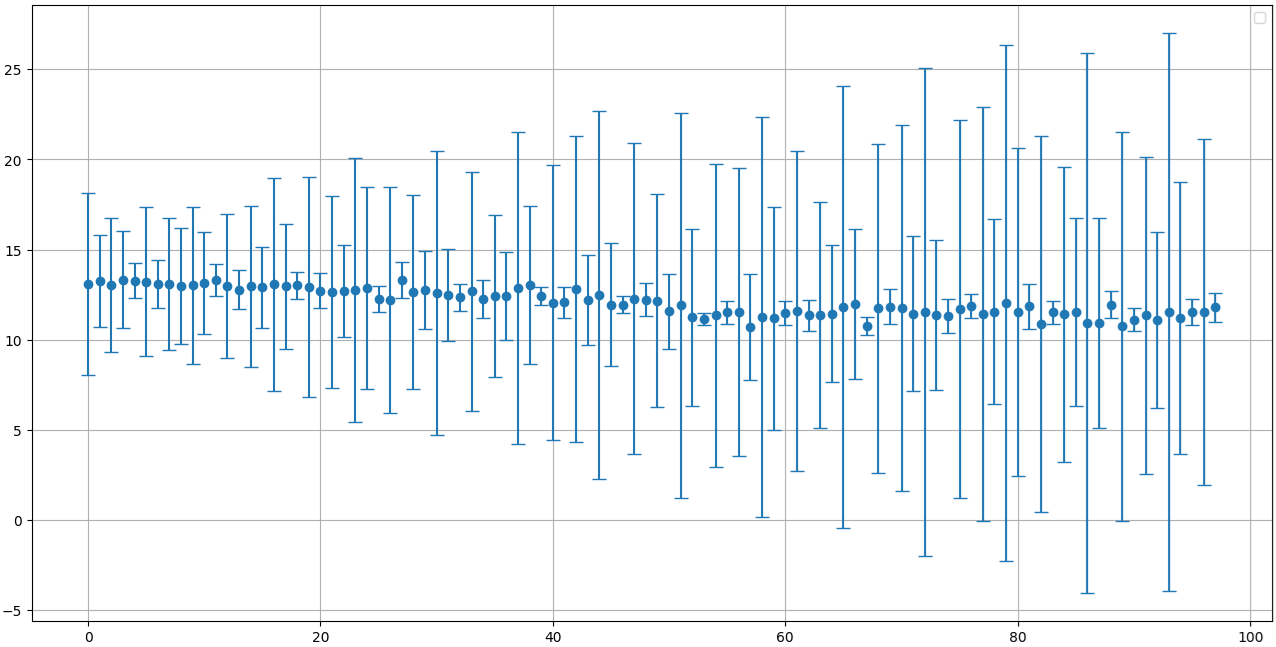}
    \caption{ResNetv2}
    \label{fig:sub3}
  \end{subfigure}
  \hfill
  \begin{subfigure}[b]{0.45\textwidth}
    \includegraphics[width=\textwidth]{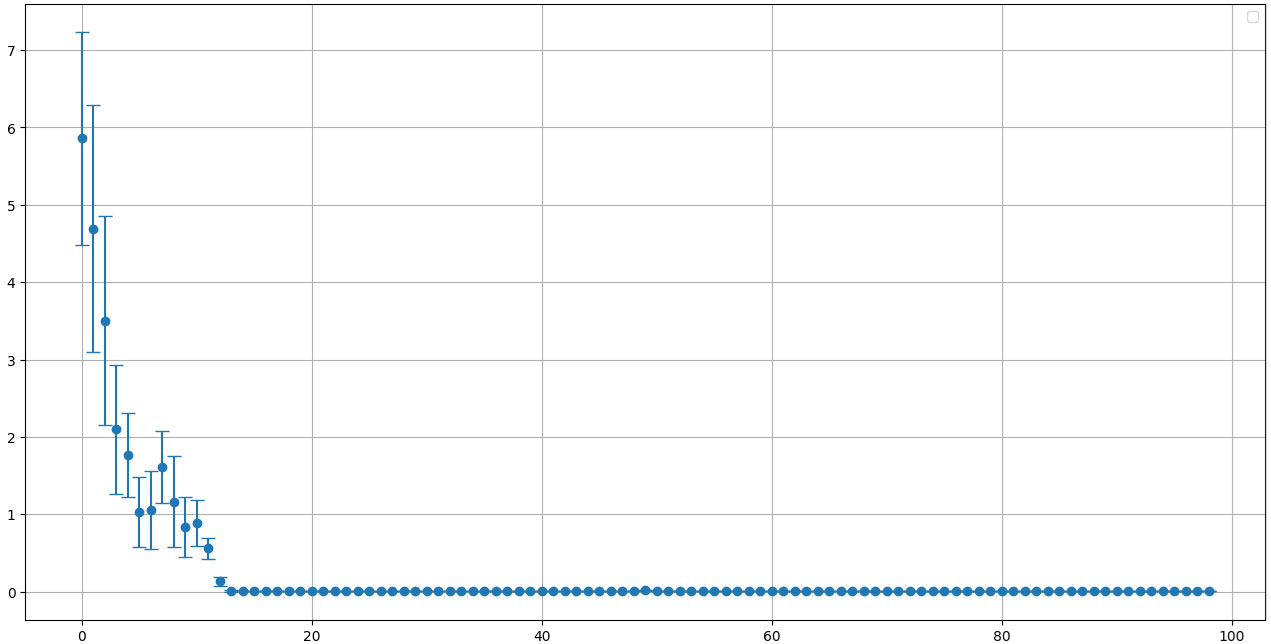}
    \caption{GloNet}
    \label{fig:sub4}
  \end{subfigure}  
  \caption{Block's outputs L1 norm, 100 blocks, 200 epochs, SGEMM. Mean and standard deviation from 5000 samples. GloNet (d) uses only the first 12 blocks, and achieves the least variance, indicating a more consistent learning of features. In ResNetv2 (c) all blocks give a similar contribution, as expected because of \bn. ResNetv2-no-BN (b) uses all outputs in different ways, because of the residual maps mixed with the skip connections. The baseline vanilla-no-BN (a) shows the opposite behavior to GloNet, emphasizing late outputs and diminishing earlier ones, indicating shorter gradient paths to the last blocks.}
  \label{fig:negligible_outputs}
\end{figure}

GloNet accumulates information in the first few blocks and uses only the required capacity for a specific task and architecture, leading to minimal output from subsequent blocks, see Remark~\ref{rem:self_regulation}. This is not observed in baseline models with or without batch normalization, see Figure~\ref{fig:negligible_outputs}, and likely contributes to GloNet's stable performance as network depth increases, in contrast to the degradation observed in the baseline models under similar depth conditions.

\textbf{MNIST Fully Connected Classification.}
To confirm that GloNet automatic choice of optimal depth and GloNet training resilience to depth were 
not associated to the particular SGEMM regression task, we performed a series of experiments with identical architecture on a completely different task: image classification with MNIST. Although using fully connected architectures for image classification is generally not the best approach, with this task we have been able to significantly increase the number of input features, which theoretically could pose a greater challenge to models that are not very deep.

\begin{figure}[tb]
\begin{subfigure}[b]{0.50\textwidth}
  \includegraphics[width=\columnwidth]{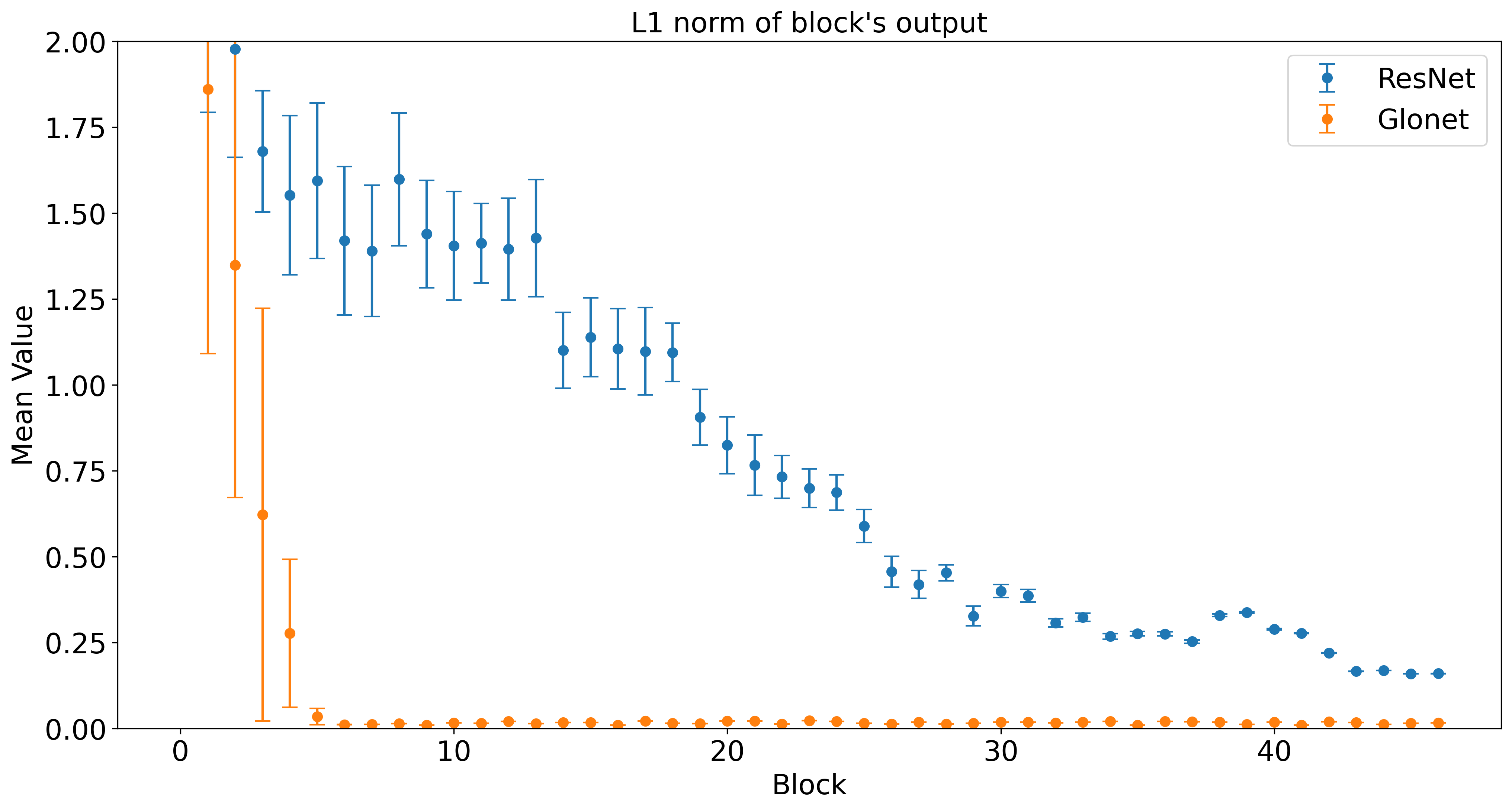}
  \caption{}
  \label{fig:negligible_outputs_mnist}
\end{subfigure}
\hfill
\begin{subfigure}[b]{0.50\textwidth}
  \includegraphics[width=\columnwidth]{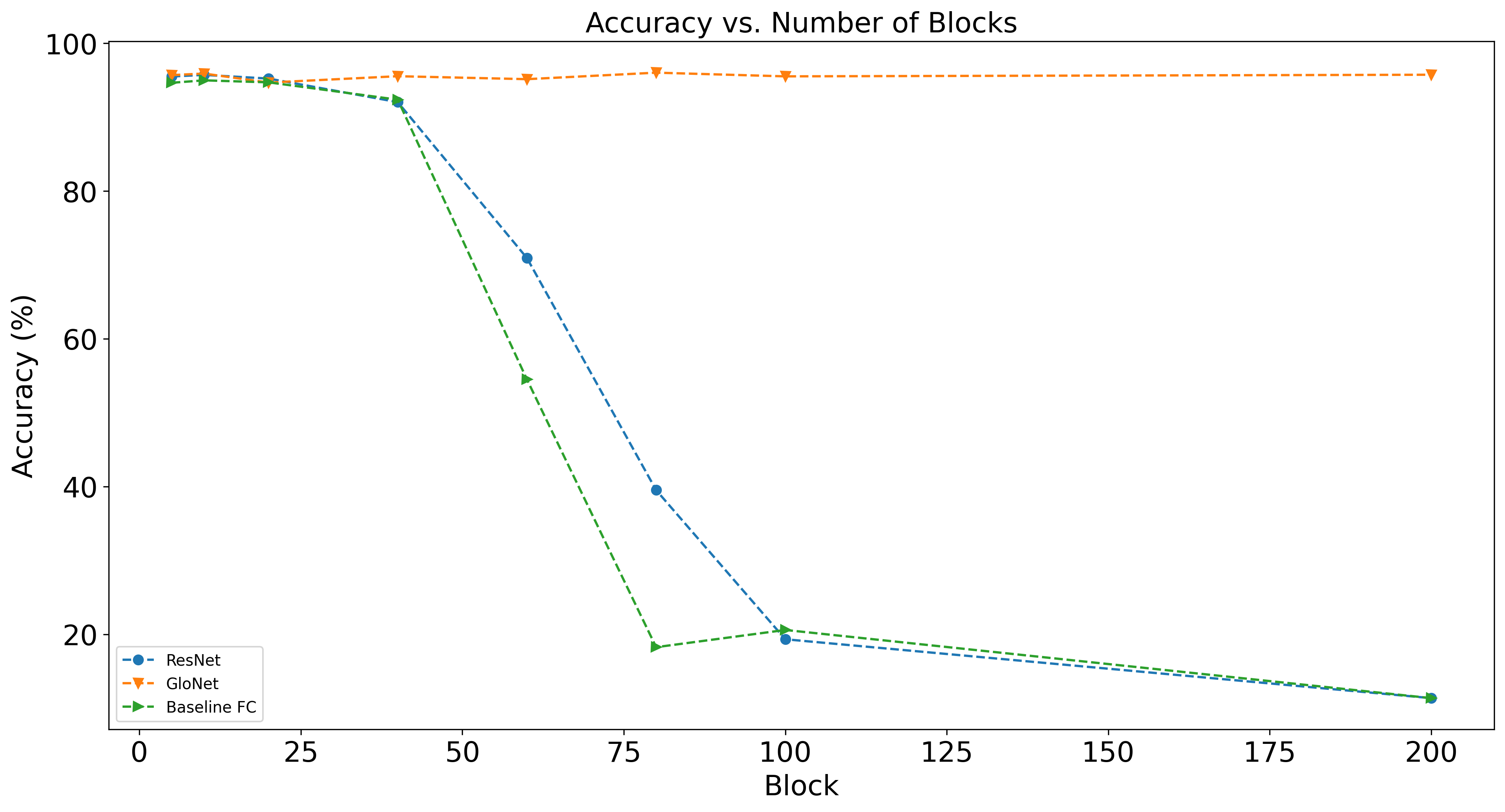}
  \caption{}
  \label{fig:deep_training_mnist}
\end{subfigure}
\caption{(\textbf{a}) Block's outputs L1 norm, GloNet (orange), and ResNetv2 (blue), 50 blocks, 200 epochs, MNIST. Mean and standard deviation from 5000 samples. GloNet uses only the first 4 blocks. In ResNetv2, all blocks give a significant contribution, as expected because of \bn. (\textbf{b}) Accuracies of GloNet (orange), ResNetv2 (blue), and vanilla (green), 200 epochs, MNIST, against $6, 10, 24, 50, 60, 80, 100$, and $200$ blocks. GloNet is consistent across depths, while vanilla and ResNetv2 show a rapid decline with increased depth.}
\end{figure}

The only difference from the architecture used in SGEMM is the head, which in this case is a fully connected layer followed by a SoftMax layer on 10 classes. We tested architectures with $6, 10, 24, 50, 60, 80, 100$, and $200$ blocks for GloNet and a convolutional baseline, halved for the corresponding ResNet baseline. 
Figure~\ref{fig:negligible_outputs_mnist} shows that GloNet automatically chooses the optimal number of blocks. However, notice that in this case, differently from Figure~\ref{fig:negligible_outputs}c, ResNetv2 outputs show a decreasing shape. This is probably due to the trainable parameters of the batch normalization, that in this case are able to force a small mean and variance on the last blocks. This indicates that also with batch normalization the network struggles to self-regulate its depth.
Figure~\ref{fig:deep_training_mnist} confirms GloNet resilience to an increasing depth, and shows a severe performance degradation of ResNetv2 when depth goes above 50 blocks.

\textbf{CIFAR10 Convolutional Classification.}
We further experimented with a ResNet20 on CIFAR10. ResNet20 is a ResNetv2 with 3 stages of 3 residual blocks each, described in~\cite{HZRDRL}. We compared a ResNet20 architecture with its GloNet version, obtained by removing the backbone and adding a GloNet layer before the classification head, as detailed in Section~\ref{sec:implementation}. 
We trained for 200 epochs. Learning curves are completely overlapping, with best test errors $91.12\%$ and $91.08\%$ for ResNet and GloNet respectively. This experiment shows that also with convolutional architectures, GloNet performs on par with the traditional ResNet architecture, despite taking half the time for training.



\begin{figure}[t!]
\begin{subfigure}[tb]{0.475\textwidth}
  \centering
  \includegraphics[width=\columnwidth]{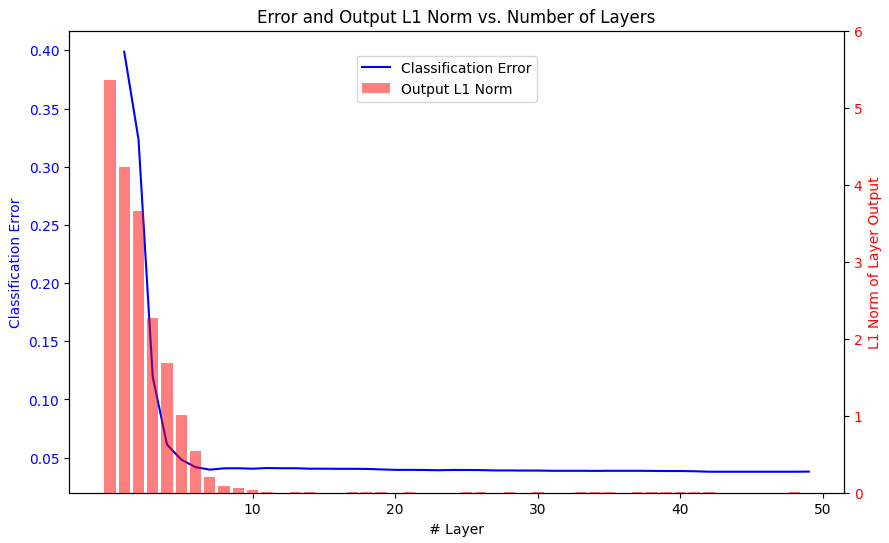}
  \caption{}
  \label{fig:removing_blocks}
\end{subfigure}
\hfill
\begin{subfigure}[ht!]{0.475\textwidth}
  \centering
  \includegraphics[width=\columnwidth]{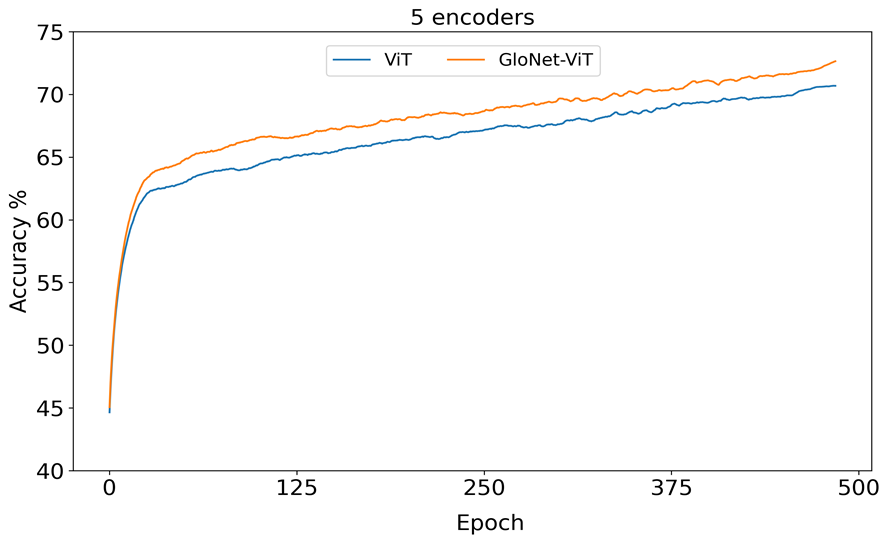}
  \caption{}
  \label{fig:vit}
\end{subfigure}
\caption{(\textbf{a}) Block's outputs L1 norm of GloNet (red), smaller networks performance (blue), 50 blocks, 200 epochs, MNIST. Performance plateaus after block eight. (\textbf{b}) Accuracy of ViT (blue), and GloNet-ViT (orange) with $5$ encoders. Best accuracies $0.709$, $0.727$ for ViT, GloNet-ViT, respectively.}
\end{figure}

\textbf{GloNet for Vision Transformer Classification.}
A Visual Transformer (ViT) is a transformer applied to sequences of feature vectors extracted from image patches~\cite{DBKIWW}. The $n$ encoders outputs can be accumulated into a GloNet layer before going to the classification head, see Section~\ref{sec:implementation} for details. In this experiment we compare ViT, with and without GloNet, on CIFAR-10, with $4, 5$ and $6$ encoders. Training plateaus at around 500 epochs, and final accuracies align with those from literature.
With 4 and 6 encoders, accuracies overlap for ViT and GloNet-ViT. With 5 encoders, GloNet-ViT appears to improve over ViT, see Figure~\ref{fig:vit}. ViT best accuracies are $0.707$, $0.709$, and $0.725$, and GloNet-ViT best accuracies are $0.709$, $0.727$, and $0.729$, for 4, 5, and 6 encoders, respectively. This is a proof-of-concept experiment showcasing the robustness and versatility of GloNet for complex architectures like transformers.

\textbf{Controllable Efficiency/Performance Trade-Off.}
In an experiment to demonstrate how GloNet can be used to choose an optimal efficiency/performance trade-off, we trained a 50-block GloNet fully connected architecture on MNIST for 200 epochs. After training, we progressively removed the last block, adjusted accordingly the GloNet layer to sum fewer blocks, and evaluated the shallower model without retraining. As shown in Figure~\ref{fig:removing_blocks}, removing up to 42 blocks did not significantly impact accuracy, illustrating GloNet's ability to balance efficiency and performance.

\section{Conclusions}
\label{sec:conclusions}

We introduce GloNet, a method designed to augment existing architectures without adding complexity or reducing performance. It effectively renders the architecture resilient to depth-related learning issues. As an alternative to ResNet, GloNet offers advantages, without any disadvantage: it achieves similar training outcomes in nearly half the time at depths where ResNet remains stable, and maintains consistent performance at greater depths where ResNet falters.

{
\small
\bibliographystyle{splncs04}
\bibliography{mau.bib}
}

\end{document}